\documentclass[sigconf]{acmart}

\usepackage{multirow}
\usepackage{float}
\usepackage{algorithm}
\usepackage{algpseudocode}
\usepackage{bbding}
\usepackage{ulem}
\usepackage{paralist}
\usepackage{tabularx} 
\AtBeginDocument{%
  }

\renewcommand\footnotetextcopyrightpermission[1]{}
\begin{document}

\title{ManipDreamer: Boosting Robotic Manipulation World Model with Action Tree and Visual Guidance}

\author{Ying Li}
\affiliation{%
  \institution{State Key Laboratory of Multimedia Information Processing, School of Computer Science, Peking University}
  \state{Beijing}
  \country{China}
}
\email{2301210309@stu.pku.edu.cn}

\author{Xiaobao Wei}
\affiliation{%
  \institution{State Key Laboratory of Multimedia Information Processing, School of Computer Science, Peking University}
  \state{Beijing}
  \country{China}
}

\author{Xiaowei Chi}
\affiliation{%
  \institution{Hong Kong University of Science and Technology}
  \state{Beijing}
  \country{China}
}
\email{litwellchi@gmail.com}

\author{Yuming Li}
\affiliation{%
  \institution{State Key Laboratory of Multimedia Information Processing, School of Computer Science, Peking University}
  \state{Beijing}
  \country{China}
}

\author{Zhongyu Zhao}
\affiliation{%
  \institution{State Key Laboratory of Multimedia Information Processing, School of Computer Science, Peking University}
  \state{Beijing}
  \country{China}
}

\author{Hao Wang}
\affiliation{%
  \institution{State Key Laboratory of Multimedia Information Processing, School of Computer Science, Peking University}
  \state{Beijing}
  \country{China}
}

\author{Ningning Ma}
\affiliation{%
  \institution{Autonomous Driving Development, NIO}
  \state{Beijing}
  \country{China}
}

\author{Ming Lu}
\affiliation{%
  \institution{State Key Laboratory of Multimedia Information Processing, School of Computer Science, Peking University}
  \state{Beijing}
  \country{China}
}

\author{Shanghang Zhang}
\affiliation{%
  \institution{State Key Laboratory of Multimedia Information Processing, School of Computer Science, Peking University}
  \state{Beijing}
  \country{China}
}

\renewcommand{\shortauthors}{Li et al.}
\settopmatter{printacmref=false}
\begin{abstract}
While recent advancements in robotic manipulation video synthesis have shown promise, significant challenges persist in ensuring effective instruction-following and achieving high visual quality.
Recent methods, like RoboDreamer, utilize linguistic decomposition to divide instructions into separate lower-level primitives. 
They condition the world model on these primitives to achieve compositional instruction-following. 
However, these separate primitives do not take into account the relationships that exist between them. 
Furthermore, recent methods neglect valuable visual guidance, including depth and semantic guidance, both crucial for enhancing visual quality.
This paper introduces ManipDreamer, an advanced world model based on the action tree and visual guidance. 
To better learn the relationships between instruction primitives, we represent the instruction as the action tree and assign embeddings to tree nodes. 
Therefore, each instruction can acquire its embeddings by navigating through the action tree. 
The instruction embeddings can be used to guide the world model. 
To enhance visual quality, we combine depth and semantic guidance by introducing a visual guidance adapter compatible with the world model. 
This visual adapter enhances both the temporal and physical consistency of video generation.
Based on the action tree and visual guidance, ManipDreamer significantly boosts the instruction-following ability and visual quality.
Comprehensive evaluations on robotic manipulation benchmarks reveal that ManipDreamer achieves large improvements in video quality metrics in both seen and unseen tasks, 
with PSNR improved from 19.55 to 21.05, SSIM improved from 0.7474 to 0.7982 and reduced Flow Error from 3.506 to 3.201 in unseen tasks, compared to the recent RoboDreamer model.
Additionally, our method increases the success rate of robotic manipulation tasks by 2.5\% in 6 RLbench tasks on average.
\end{abstract}




\received{20 February 2007}
\received[revised]{12 March 2009}
\received[accepted]{5 June 2009}

\maketitle

\vspace{-2mm}
\section{Introduction}
\label{sec:intro}

Video generation technologies have developed into three main technical paradigms based on control modalities: text-to-video (T2V), image/video-to-video (I2V), and other modalities to video.
Early works focus on generating video content that is conditioned by textual input (T2V)~\cite{pan2017create,marwah2017attentive,li2018video,gupta2018imagine,liu2019cross,hong2022cogvideo,ho2022video,singer2022make}. 
Recent advancements in I2V~\cite{xing2024dynamicrafter,peng2025open,wang2025wan} and V2V~\cite{agarwal2025cosmos} techniques are focused on generating high-fidelity videos with complex motions and interactions. These developments illustrate the potential of using video generative models as world models. 
Other modalities of video generation methods have been proposed, 
such as motion trajectory-controlled approaches~\cite{wang2024motionctrl,yin2023dragnuwa} that enable precise perspective movement,
depth-guided synthesis~\cite{he2023animate,xing2024make} that enhances realism and spatial coherence, 
and pose-driven~\cite{karras2023dreampose, hu2024animate, wei2024gazegaussian, wei2024graphavatar, chen2024mixedgaussianavatar, chen2025diffusiontalker} methods that generate human animation videos. 

Creating high-quality datasets is essential for data-intensive fields, such as embodied intelligence~\cite{zhou2024robodreamer, wang2024language, gaoflip, chi2024eva} and autonomous driving~\cite{ni2025maskgwm, russell2025gaia, zhao2024drivedreamer4d, huang2024S3G, wei2024emd}, where collecting physical data is often prohibitively expensive. 
Recent developments in generative models have enhanced the creation of world models, providing a cost-effective solution for generating synthetic data. 
In autonomous driving, world models have shown impressive versatility by integrating various control signals to create physically consistent scenarios. 
Many works~\cite{gao2024vista, gao2023magicdrive, ni2025maskgwm, russell2025gaia, zhao2024drivedreamer4d, lu2024drivingrecon} utilize control signals such as velocity profiles, steering angles, and traffic rule embeddings to condition the generation process, which is essential for applications like safety validation and rare-scenario augmentation. 
Methods~\cite{ni2025maskgwm} based on diffusion transformers leverage MAE-style feature learning to generate multi-view driving videos based on driving actions and text prompts. 
Large-scale models~\cite{russell2025gaia} integrate LiDAR point clouds, BEV semantic maps, and traffic agent trajectories to generate geographically diverse multi-camera streams, facilitating closed-loop simulations for planning algorithms. 
4D scene reconstruction~\cite{zhao2024drivedreamer4d} utilizes world model priors to synthesize novel trajectory videos, showcasing the synchronized spatial-temporal evolution of dynamic objects and static environments. 

In contrast, robotic video generation focuses on task-oriented synthesis guided by specific inductive biases.
Typically, a robotic world model takes a scene snapshot and generates robotic videos conditioned on language instructions.
Unlike approaches in autonomous driving, most existing methods rely solely on the RGB modality, which provides limited multi-modal information. 
Robodreamer~\cite{zhou2024robodreamer} breaks down language instructions into lower-level primitives for the synthesis of compositional manipulation. Wang et al.~\cite{wang2024language} combine gesture-based spatial references with deictic language to clarify ambiguities in open-world environments. 
For action-conditioned generation, 
FLIP~\cite{gaoflip} introduces flow-centric planning, where optical flow fields act as intermediate representations of actions to guide video generation. 
Recent efforts, such as EVA~\cite{chi2024eva}, have introduced an Understand–Imagine–Predict world model loop.
However, it still relies solely on RGB images as the input perception modality, limiting its ability to capture richer spatial and semantic context. 
By overcoming these limitations, world models can achieve a more accurate understanding of spatial affordances and physical constraints, ultimately enhancing their performance in sim-to-real transfer and complex manipulation tasks.

While recent advancements in robotic world modeling show promise, significant challenges remain in ensuring effective instruction following and achieving high visual quality. 
This paper proposes ManipDreamer, an advanced robotic manipulation world model based on the action tree and multi-modal visual guidance. 
For instruction following, instead of breaking down language instructions into lower-level primitives, 
we represent each instruction as an action tree. 
We assign embeddings to the nodes of the action tree, allowing each instruction to derive its own embeddings through a traversal of the tree. 
These instruction embeddings are then utilized to guide the world model. 

Regarding visual quality, we observe that generated robotic videos exhibit significant artifacts. These issues include structural anomalies in the robotic arms, geometric deformations of objects, and temporal inconsistencies such as cross-frame attribute mutations. 
As a result, the fidelity of the synthetic content is compromised, affecting both spatial continuity and temporal coherence. 
We introduce a visual adapter to leverage the depth and semantic information in the video generation process to enhance the visual quality. 
The adapter features a multi-scale encoder that progressively incorporates depth-aware geometric constraints into the video diffusion pipeline.

Our experiments demonstrate that ManipDreamer significantly improves the performance of robotic manipulation world models. 
Specifically, it achieves notable enhancements in video quality metrics such as PSNR, SSIM, and Flow Error, surpassing the recent RoboDreamer model. 
Additionally, our method shows consistent improvements in the success rate of robotic manipulation tasks across multiple RLbench\cite{james2020rlbench} benchmarks. 
Our main contributions include: 
\begin{compactitem}
\item We propose an advanced world model ManipDreamer for robotic manipulation that ensures effective instruction-following and achieves high visual quality. 
\item We propose representing the instruction as an action tree to learn the relationships between instruction primitives.
\item We enhance visual quality with a well-designed adapter that integrates depth and semantics into robotic world models through hierarchical guidance. 
\item We conduct a comprehensive experiment to evaluate the effectiveness of ManipDreamer, achieving significant improvements in both video quality and task compliance across six benchmark datasets. 
\end{compactitem}

\vspace{-4mm}
\section{Related Works}
\label{sec:related}

\vspace{-1mm}
    \subsection{Robotic Video Generation}
    \label{robotic_video_generation}
    Robotic video generation differs fundamentally from general video synthesis tasks in that it demands task-driven reasoning, physical interaction modeling, and scene-level spatial understanding.
    ManiGaussian~\cite{lu2024manigaussian} employs dynamic Gaussian splatting on depth-derived point clouds for policy video generation but encounters rendering artifacts due to the Gaussian representation. 
    RoboAgent~\cite{bharadhwaj2024roboagent} leverages SAM-2~\cite{ravi2024sam} masks for data augmentation, rather than using semantic features as direct generative guidance. 
    Recent advances in diffusion models have led to promising progress in the world model for robotics. 
    AVDC~\cite{ko2023learning} pioneers action-conditioned video generation via latent diffusion without requiring explicit action labels, demonstrating the potential of synthetic data for downstream policy learning. 
    RoboDreamer~\cite{zhou2024robodreamer} improves instruction generalization by decomposing language commands into reusable sub-task components. 
    This\&That~\cite{wang2024language} addresses language ambiguity by grounding instructions in both textual and gestural modalities. 
    FLIP~\cite{gaoflip} proposes a flow-centric world model for manipulation planning, unifying flow prediction, flow-conditioned video generation, and vision-language value estimation.
    
    Despite these advancements, most existing methods rely solely on RGB frames and textual prompts, without explicitly modeling scene geometry or object-level semantics.
    This limits their ability to preserve spatial consistency and to capture fine-grained manipulation constraints, both of which are critical for generating realistic and executable robot videos in complex environments. Furthermore, these approaches predominantly focus on a single modality of scene representation, restricting their capacity to fully comprehend and reproduce the rich, multi-faceted nature of robotic interactions in real-world settings.

\begin{figure*}[!ht]
\centering
\includegraphics[width=\textwidth]{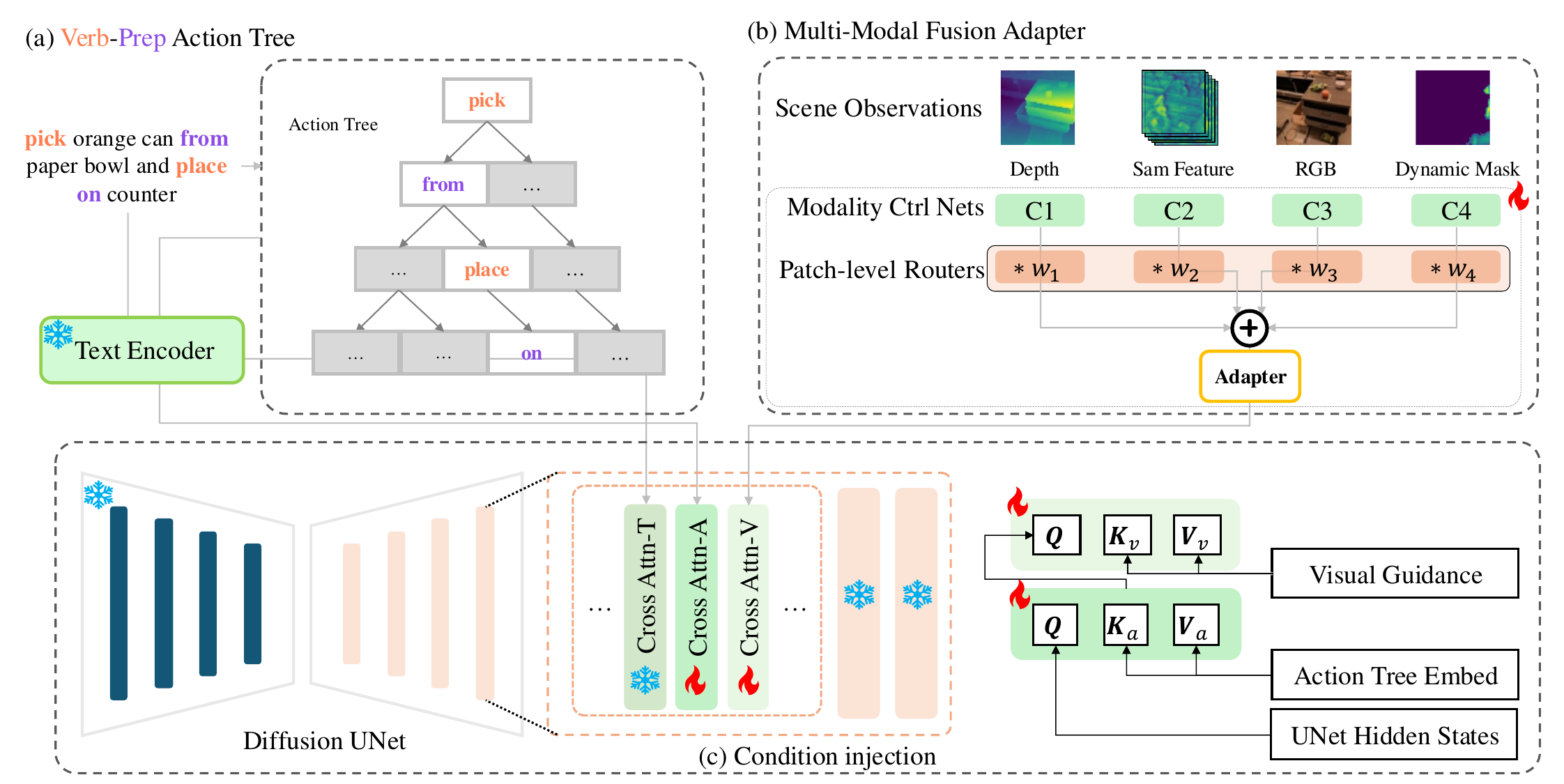}
\vspace{-2mm}
\caption{Overview of ManipDreamer: a world model for robotic manipulation that integrates structured instruction representations (a) and multi-modal visual guidance  (b).  
We encode language instructions as verb-preposition action trees to capture compositional task structure and inject depth and semantic features through a hierarchical adapter to enhance spatial-temporal consistency in video generation. In the UNet decoder, action tree embeddings and visual guidance features are sequentially injected every 3 layers using cross-attention mechanisms.(c)}
\label{fig:method_overview}
\vspace{-2mm}
\end{figure*}

    \subsection{Multi-modal Conditioned Generation}
    \label{multi_modal-conditioned-generation}
    ControlNet~\cite{zhang2023adding} revolutionized conditional image generation by enabling spatial constraints while preserving the capabilities of the original model. 
    Subsequent works have extended this idea to the video domain. For example, SparseCtrl~\cite{guo2024sparsectrl} introduces sparse frame guidance to enhance temporal consistency,
    and Ctrl-Adapter~\cite{lin2024ctrl} adapts pretrained ControlNets to various video generation backbones (e.g., I2VGen-XL~\cite{zhang2023i2vgen}) through parameter-efficient fine-tuning. 
    These methods demonstrate strong performance in incorporating diverse modalities to guide video generation. 
    However, they typically rely on sparse but explicit multi-frame conditions, and are therefore ill-suited for scenarios like robotic video generation, 
    where only the initial frame and a textual prompt are available at inference time. 
    These approaches primarily focus on frame-to-frame conditioning patterns, where conditioned frames directly control corresponding output frames. This proves challenging in robotic world model scenarios that require long-horizon reasoning from a single initial observation, necessitating alternative conditioning mechanisms for effective trajectory synthesis.
    
    Contrary to the aforementioned methods focused on natural image generation, recent autonomous driving (AD) approaches have begun to exploit scene structure through geometric representations.
    SyntheOCC~\cite{li2024syntheocc} combines multi-plane images with semantic occupancy maps to construct realistic driving environments,
    while UniScene~\cite{li2024uniscene} synthesizes multi-view videos using BEV-derived LiDAR and occupancy features. InfiniCube~\cite{lu2024infinicube} introduces a scalable world-guided framework for generating unbounded dynamic 3D driving scenes. 
    DriveDreamer4D~\cite{zhao2024drivedreamer4d} and ReconDreamer~\cite{ni2024recondreamer} introduce a 4D driving scene generation framework that leverages world model priors to synthesize videos. 
    The introduction of these novel modalities has significantly enhanced scene understanding and motion prediction capabilities, demonstrating that incorporating geometric and semantic representations can substantially improve the quality and consistency of generated videos. This inspires our approach to leverage complementary modalities for more effective robotic scene understanding and manipulation planning.
    
    However, these methods primarily target high-fidelity video generation in structured scenarios with simple road trajectories. 
    In contrast, robotic video generation typically involves decision-making and interaction within complex, unstructured environments, where spatial understanding and semantic awareness are essential.
    To address this challenge, we propose a novel framework that synergistically integrates multi-modal cues into robotic world models via hierarchical guidance.

    \subsection{Action Tree in Robotics}
    Behavior Trees (BTs) have been widely adopted in traditional robotic manipulation as a modular and interpretable framework for task execution~\cite{ghzouli2020behavior}. 
    They offer a hierarchical and extensible structure for decomposing complex tasks into atomic behaviors (e.g., \textit{grasp}, \textit{move}) and control operators (e.g., \textit{sequence}, \textit{fallback}). 
    This structured representation supports reusability, debugging, and runtime adaptability, making BTs well-suited for modeling reactive behaviors in dynamic environments~\cite{ghzouli2020behavior}. 
    Recent studies further show that BTs align with the models-at-runtime paradigm, enabling seamless coordination of low-level robot skills through high-level task specifications. 
    FOON~\cite{paulius2016functional} proposes a structured bipartite graph representation, enabling robots to retrieve task trees and generate manipulation trajectories. 
    Michele et al.~\cite{colledanchise2021implementation} provide a practical analysis of Behavior Tree implementations in robotics, detailing execution engines, memory nodes, node parameterization, and asynchronous action handling for real-world robot control systems. 
    Approximate Task Tree Retrieval~\cite{sakib2022approximate} proposes a method for generating executable task trees from a knowledge graph. 
    Corrado et al.~\cite{pezzato2023active} proposes a hybrid framework combining active inference with behavior trees, enabling reactive robotic planning and execution. 
    Long-Horizon FOON~\cite{paulius2023long} proposes a hierarchical task planning framework that transforms symbolic FOON representations into executable robot plans. 
    Yang et al.~\cite{yang2014manipulation} represent manipulation actions as treebanks encoding physical constraints and task dependencies for action retrieval and adaptation. 
    Another work~\cite{yang2023robot} further bridges language and BTs by using language models (LMs) to generate XML-formatted trees from natural instructions.

    Unlike prior works that leverage action trees primarily for low-level control or symbolic planning,
    our approach is the first to integrate action tree semantics into generative models for robotic video synthesis.
    By embedding structured task representations into a diffusion-based video generation pipeline,
    we enforce spatiotemporal consistency between complex language instructions (e.g., “pick up the cup and pour water”) and the resulting motions.
    This demonstrates that action trees serve as a crucial symbolic prior to constrain the open-ended space of generative robotics.
\section{Method}
\label{sec:method}
In this section, we first review the preliminaries of robotic video generation using world models in Sec.~\ref{subsec:problem_formulation}.
We then introduce our proposed world model, ManipDreamer, which aims to achieve accurate instruction-following and high-fidelity visual synthesis.
To capture the relationships among instruction primitives, we present the Verb-Preposition Action Tree in Sec.~\ref{subsec:action_tree_conditioning}.
Finally, to improve the visual quality of generated robotic videos, we integrate Multi-modal Visual Guidance into the world model, as detailed in Sec.~\ref{subsec:multi-modal_control_adapter}.

\subsection{Problem Formulation}
\label{subsec:problem_formulation}
  A typical robotic world model includes a video generation module, enabling interaction through video outputs to convey reasoning results and enhance interpretability.
  Given an initial scene observation $I \in \mathbb{R}^{3\times H\times W}$ and text instruction $c_{\text{text}}$, 
  a robotic video generator aims to produce a physically plausible trajectory $V\in \mathbb{R}^{F\times 3\times H\times W}$
  depicting $F$ subsequent frames of environment evolution under specified manipulation. 
  Traditional approaches employ diffusion models $\mathcal{F}_\theta (z_t, t, c_{\text{text}})$
  that gradually denoise latent representations $z_t$
  through iterative prediction of noise components $\epsilon_\theta(z_t, t, c_{\text{text}})$
  conditioned on time $t$ and $c_{\text{text}}$ minimizing the loss:
\begin{equation}
\mathcal{L}_{\text{diff}} = \mathbb{E}_{z,\epsilon \sim \mathcal{N}(0,I),\, t} \left\| \epsilon - \epsilon_\theta(z_t, t, c_{\text{text}}) \right\|_2^2,
\end{equation}
  Our extended formulation incorporates geometric depth $D_0\in \mathbb{R}^{1\times H\times W}$, 
  semantic segmentation $S_0\in \mathbb{R}^{C_{\text{sam}}\times H\times W}$, 
  and dynamic mask $M_0\in \mathbb{R}^{1\times H\times W}$, 
  all tied to the first frame $I_0$. 
  These inputs help the enhanced model $\mathcal{F}_\theta (z_t, t, c_{\text{text}}, S_0, D_0)$ 
  achieve stronger spatial understanding and execute more accurate manipulations, 
  while preserving object identity and allowing flexible state changes.

\begin{figure}[!ht]
\centering
\includegraphics[width=1.0\linewidth]{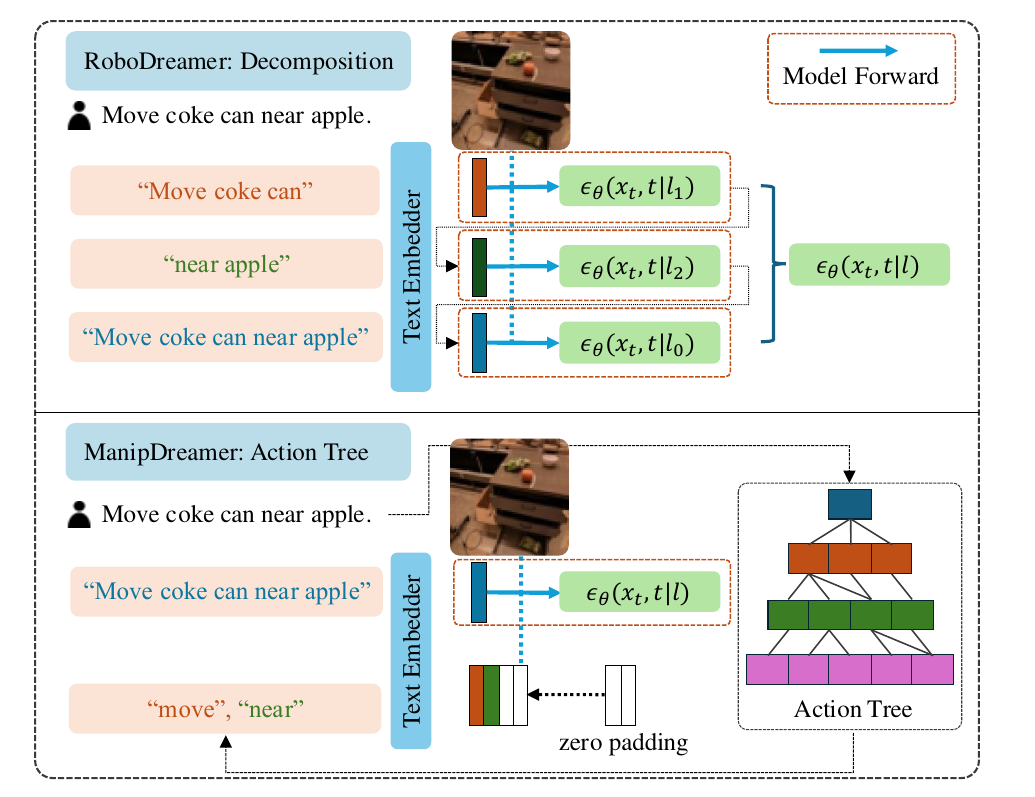}
\caption{Different from the instruction decomposition manner in RoboDreamer, we propose a novel action tree method to represent the action to be generated consuming less computation resource.}
\label{fig:action_tree}
\end{figure}

\subsection{Verb-Preposition Action Tree}
\label{subsec:action_tree_conditioning}
  \subsubsection{Action Tree Construction}
  \label{subsubsec:action_tree_construction}
RoboDreamer~\cite{zhou2024robodreamer} decomposes language instructions into sub-actions to improve generalization.
However, its decomposition-and-composition strategy suffers from several limitations: 
\begin{itemize} 
    \item The decomposition ignores the temporal dependencies between sub-actions by using the average of predicted noises; 
    \item Executing the model separately for each sub-action incurs significant GPU memory and computational cost, especially for long instructions; 
    \item The denoising trajectories of individual sub-actions may be inconsistent with the overall video generation process. 
\end{itemize}

We propose a tree-based structure as an alternative mechanism to control the action generation process.
The key difference between our method and RoboDreamer is illustrated in Fig.~\ref{fig:action_tree}.
To reduce the solution space, we represent each instruction as a hierarchical action tree composed of reusable embeddings of key action words—namely, verbs (e.g., pick, place) and prepositions (e.g., on, in).
For each language instruction in the dataset, we extract its constituent action words and assign them to the appropriate layers in the tree. The action words are in order from the root of the tree down to the leaves.
Finally we get a action tree tree with alternating layers of verbs and prepositionsas shown in Fig. \ref{fig:method_overview}. Each key action word is then embedded using the CLIP text encoder, which is also used in the original text-conditioning pathway. This tree structure aligns well with the compositional and hierarchical structure of complex actions in possible language instructions.

\subsubsection{Action Tree Parsing}
To generate action condition for a video generation model, we obtain the text embeddings on the path nodes after traveling down the action tree. For example, we get the text embedding of words "pick", "from", "place", and "in" after traveling the instruction "pick the apple from the table and place it in the top drawer" down the constructed action tree.
These embeddings are afterwards concatenated together to form the final action tree embedding. We fix the number of embeddings to the maximum number of action words in all possible text instructions, and zero embeddings are used for padding if the number of key words are less than that when encountering short instructions. 
Giving an instruction $c_{\text{text}}$, the final action tree embedding is defined as:
\begin{equation}
\mathcal{T}(c_{\text{text}}) = \bigoplus_{i=1}^n \big(\text{CLIP-Embed}(\text{Verb}_i) \bigoplus \text{CLIP-Embed}(\text{Prep}_i)\big),
\end{equation}
 in which $Verb_i$ and $Prep_i$ are the two adjacent verb and preposition action words. 

By encoding the full sequence of verbs and prepositions in a unified embedding, our method addresses the aforementioned key limitations of prior decomposition-based approaches. Our model captures the temporal dependencies between sub-actions as we view the actions as a whole and extract the most important part of it.
The one forward property of the key action embedding avoids extra computational overhead.
Moreover, generating actions with coherent tree embeddings ensures that the denoising trajectory remains globally consistent throughout the video generation process.

\subsection{Multi-modal Visual Guidance}
\label{subsec:multi-modal_control_adapter}
  \subsubsection{Multi Scene Modality Acquisition}
  \label{subsubsec:multi-modal_feature_extraction}
  
    Due to the lack and low quality of depth and semantic annotations in third-view robotic datasets, we leverage pretrained foundation models to estimate these modalities without the need for costly manual labeling.
    Specifically, the Depth Anything v2 model~\cite{yang2025depth} provides view-consistent and accurate depth estimation using a transformer-based monocular predictor. 
    For semantic guidance, we adopt SAM-2~\cite{ravi2024sam}, which produces semantic embeddings that support instance-aware and multi-granularity object segmentation. 
    To better preserve the original model’s representation capacity and avoid object-class misalignment, we use normalized depth maps as depth conditions and semantic embeddings—instead of discrete segmentation maps—as semantic conditions.
    To obtain the dynamic mask, we compute the similarity between SAM features from the first frame and subsequent frames, identifying regions with significant changes as potential dynamic parts.
    We also experiment with DINO~\cite{oquab2023dinov2} features and observe comparable results.
    Details of the dynamic mask algorithm are provided in the supplementary material. 
    In addition to depth, semantics, and dynamic mask, we also incorporate the RGB image as a complementary modality.
    As a result, our model is conditioned on four visual modalities: depth, semantic embedding, RGB, and dynamic mask. 

  \subsubsection{Hierarchical Feature Extraction with ControlNets}
  \label{subsubsec:hierarchical_feature_extraction}
  
    As illustrated in Fig.~\ref{fig:method_overview}, each visual modality is processed through a dedicated ControlNet branch that extracts pyramid features. 
    Commonly, ControlNet initializes its input/output convolution layers with zeros to avoid perturbing the pretrained UNet features, while inheriting encoder weights from the base UNet backbone.
    However, we find that the original ControlNet design is suboptimal for multi-modal guidance in our setting, for the following reasons:
    First, each ControlNet receives a single-frame input, and replicating this frame across all $T$ time steps leads to significant GPU memory overhead and increased inference latency.
    Second, when repeatedly feeding the same frame to a 3D ControlNet, we observe that its outputs remain nearly identical across time, suggesting limited temporal contribution. 
    To address these issues, we propose a 2D ControlNet structure, which inherits the temporal-agnostic parameters from the base UNet and replaces 3D convolutions with 2D convolutions.
    This modification preserves spatial guidance while greatly improving computational efficiency for frame-wise visual conditions.

  Formally, given frame-replicated control signals 
  $D\in \mathbb{R}^{C_{\text{depth}} \times H \times W}$,
  $S\in \mathbb{R}^{C_{\text{sam}} \times H \times W}$, 
  $R\in \mathbb{R}^{C_{\text{rgb}} \times H \times W}$, 
  and $M\in \mathbb{R}^{C_{\text{mask}}\times H\times W}$, 
  where $C_{\text{depth}}$, $C_{\text{sam}}$, $C_{\text{rgb}}$, and $C_{\text{mask}}$ are the channel numbers of depth, semantic, RGB, and dynamic mask features respectively,
  we compute pyramidal features $\mathcal{P}_d$, $\mathcal{P}_s$, $\mathcal{P}_r$, and $\mathcal{P}_m$ through the ControlNet branches. The above process can be formulated as: 
  \begin{equation}
  \left\{
  \begin{array}{ll}
      \mathcal{P}_d = \text{ControlNet}_{\text{depth}}(D) \\
      \mathcal{P}_s = \text{ControlNet}_{\text{sam}}(S) \\
      \mathcal{P}_r = \text{ControlNet}_{\text{rgb}}(R) \\
      \mathcal{P}_m = \text{ControlNet}_{\text{mask}}(M)
  \end{array},
  \right.
  \end{equation}

  To fuse each layer of features from different modalities,
  we use a patch-lever q-former router \cite{lin2024ctrl} $\Phi$ that takes the concatenated features from different modalities of this layer as input,
  and outputs the score $\varphi$ for each modality in a patch-lever manner:
  \begin{align}
    \mathbf{\varphi^{(l)}[k]} &= \Phi\Big( p_d^{(l)}[k], p_s^{(l)}[k], p_r^{(l)}[k], p_m^{(l)}[k] \Big),
  \end{align}
  Then the features are multiplied by the scores and added together to get the fused feature of this layer: 
  \begin{align}
    \mathcal{P}_{\text{fused}}^{(l)}[k] &= \mathbf{\varphi^{(l)}[k]} \odot \Big( p_d^{(l)}[k], p_s^{(l)}[k], p_r^{(l)}[k], p_m^{(l)}[k] \Big),
  \end{align}
  where $l$ is the pyramid level, $k$ is the patch index, $\odot$ denotes element-wise multiplication and $\mathcal{P}_{\text{fused}}$ is the fused feature.

  \subsubsection{Spatial Feature Adapters}
  \label{subsubsec:spatial_feature_adapters}
  
  To form the final control signal for UNet, we first repeat the spatial fused features $\mathcal{P}_{\text{fused}}$ 
  along the temporal dimension to get $\mathcal{P}'_{\text{fused}} \in \mathbb{R}^{T\times C_{\text{fused}}\times H\times W}$,
  Then we use a spatial-temporal feature adapter to adapt the fused features before feeding them into the base UNet. 
  Our fusion module employs sequential spatial and temporal convolutions 
  followed by spatial and temporal cross-attention layers inspired by~\cite{lin2024ctrl}.
  In the temporal attention layer, the frames are added with temporal positional embeddings to introduce extra temporal difference,
  so that the model can learn more temporal-aware features.
  The four submodules in the feature adapter are:
\begin{equation}
\left\{
\begin{array}{l}
    \mathcal{F}_{1} = \text{Conv}_{\text{spatial}}(\mathcal{P}'_{\text{fused}}) \\
    \mathcal{F}_{2} = \text{Conv}_{\text{temporal}}(\mathcal{F}_{1}) \\
    \mathcal{F}_{3} = \text{Attn}_{\text{spatial}}(\mathcal{F}_{2}) \\
    \mathcal{G} = \text{Attn}_{\text{temporal}}(\mathcal{F}_{3})\\
\end{array},
\right.
\label{eq:adapter_pipeline}
\end{equation}

To effectively integrate the final guidance features $\mathcal{G}$ with the corresponding UNet decoder blocks at each layer,
we compare two fusion strategies:
\noindent\textbf{Additive Fusion.} At each layer, the guidance features from $\mathcal{G}$ are directly added to the residual features from $\mathcal{F}_{\text{unet}}$ of the corresponding UNet encoder.
\noindent\textbf{Cross-Attention Fusion.} At each decoder block, we apply cross-attention where the query is the decoder output, and the key and value are from the guidance features $\mathcal{G}$. The impact of these two fusion strategies is evaluated in the ablation study presented in the experimental section.
By introducing pyramid features, our model enables multi-level geometric and semantic conditioning throughout the denoising process. 
It's worth noting that, in contrast to Ctrl-Adapter~\cite{lin2024ctrl}, which requires sparse, per-frame control signals, our framework relies only on single-frame inputs, making it more efficient and better suited for robotic video generation scenarios.
\begin{table*}[htbp]
    \centering
    \caption{Comparison of Action Tree vs. Linguistic Decomposition in RoboDreamer~\cite{zhou2024robodreamer} on Video Quality Metrics. The better results are in bold.}
    \label{tab:action_tree_exp}
    \resizebox{\linewidth}{!}{
        \begin{tabular}{c|ccccc|ccccc}
            \toprule
            \multirow{2}{*}{\textbf{Method}}&\multicolumn{5}{c|}{\textbf{Seen Tasks}} & \multicolumn{5}{c}{\textbf{Unseen Tasks}} \\
            & \textbf{FID} $\downarrow$ & \textbf{SSIM} $\uparrow$ & \textbf{PSNR} $\uparrow$ & \textbf{LPIPS} $\downarrow$ & \textbf{Flow Error} $\downarrow$ 
             & \textbf{FID} $\downarrow$ & \textbf{SSIM} $\uparrow$ & \textbf{PSNR} $\uparrow$ & \textbf{LPIPS} $\downarrow$ & \textbf{Flow Error} $\downarrow$ \\
            \midrule
            
            RoboDreamer (w/ linguistic decomposition)
            & 7.014 & 0.7235 & 19.47 & 0.07484 & 4.064
            & \textbf{7.336} & 0.7390 & 19.33 & 0.07634 & 3.885 \\
            RoboDreamer (w/ action tree)
            & \textbf{6.252} & \textbf{0.7358} & \textbf{19.79} & \textbf{0.07120} & \textbf{3.723}
            & 7.433 & \textbf{0.7461} & \textbf{19.50} & \textbf{0.07413} & \textbf{3.652} \\
        \bottomrule
    \end{tabular}
}
\end{table*}

\begin{table*}[htbp]
    \centering
    \caption{Comparison of different guidance methods on video quality metrics towards RoboDreamer. The best results of all are in bold and the best result of ControlNet variant are underlined. Semantic stands for SAM feature and Mask stands for dynamic mask.}
    \vspace{-2mm}
    \label{tab:video_quality}
    \resizebox{\linewidth}{!}{
        \begin{tabular}{c|c|cccc|ccccc|ccccc}
            \toprule
            \multirow{2}{*}{\textbf{Method}} & \multirow{2}{*}{\textbf{Action Tree}}&\multicolumn{4}{c|}{\textbf{Visual Input}}&\multicolumn{5}{c|}{\textbf{Seen Tasks}} & \multicolumn{5}{c}{\textbf{Unseen Tasks}} \\
            &&\textbf{Depth}&\textbf{Semantic}&\textbf{RGB}&\textbf{Mask}
             & \textbf{FID} $\downarrow$ & \textbf{SSIM} $\uparrow$ & \textbf{PSNR} $\uparrow$ & \textbf{LPIPS} $\downarrow$ & \textbf{Flow Error} $\downarrow$ 
             & \textbf{FID} $\downarrow$ & \textbf{SSIM} $\uparrow$ & \textbf{PSNR} $\uparrow$ & \textbf{LPIPS} $\downarrow$ & \textbf{Flow Error} $\downarrow$ \\
            \midrule
            Vanilla
            &\XSolid&\XSolid&\XSolid&\Checkmark&\XSolid
            & \textbf{6.0721} & 0.7300 & 19.66 & 0.07470 & 3.8004
            & \textbf{7.396} & 0.7474 & 19.5474 & 0.07478 & 3.506 \\
            \midrule
            \multirow{4}{*}{ControlNets}
            &\XSolid&\Checkmark&\XSolid&\XSolid&\XSolid
            & \underline{6.378} & 0.7355 & 19.75 & 0.07338 & 3.611
            & \underline{8.537} & 0.7543 & 19.69 & 0.07276 & 3.318 \\
            &\XSolid&\XSolid&\Checkmark&\XSolid&\XSolid
            & 6.563 & \underline{0.7375} & 19.80 & \underline{0.07265} & 3.555
            & 8.696 & \underline{0.7568} & \underline{19.75} & \underline{0.07259} & 3.265 \\
            &\XSolid&\XSolid&\XSolid&\Checkmark&\XSolid
            & 6.600 & 0.7374 & \underline{19.81} & 0.07267 & \underline{3.545}
            & 8.668 & 0.7564 & \underline{19.75} & 0.07267 & \underline{3.257} \\
            &\XSolid&\XSolid&\XSolid&\XSolid&\Checkmark
            & 6.456 & 0.7369 & 19.79 & 0.07288 & 3.566
            & 11.177 & 0.7528 & 19.67 & 0.07490 & 3.348 \\
            \midrule
            \multirow{3}{*}{ManipDreamer}
            &\Checkmark&\XSolid&\XSolid&\XSolid&\XSolid
            & 6.252 & 0.7358 & 19.79 & 0.07120 & 3.723
            & 7.433 & 0.7461 & 19.50 & 0.07413 & 3.652 \\
            &\XSolid&\Checkmark&\Checkmark&\Checkmark&\Checkmark
            & 8.182 & 0.7853 & 21.06 & 0.05702 &  \textbf{3.049}
            & 9.480 & \textbf{0.7982} & \textbf{21.05} & \textbf{0.05768} & \textbf{3.021}\\
            &\Checkmark&\Checkmark&\Checkmark&\Checkmark&\Checkmark
            & 7.209 & \textbf{0.7867} & \textbf{21.13} & \textbf{0.05460} & 3.145
            & 8.483 & 0.7949 & 20.98 & 0.05782 & 3.178 \\
            \bottomrule
        \end{tabular}
    }
    \vspace{-2mm}
\end{table*}

\section{Experiments}
\label{sec:experiments}

    We structure our experimental analysis as follows: Section~\ref{subsec:datasets_and_metrics} details the dataset configuration and evaluation protocols. Section~\ref{subsec:results} presents comprehensive comparisons with our baseline methods, while Section~\ref{subsec:ablation_study} systematically examines several interesting questions.

    \subsection{Experimental Setup}
    \label{subsec:datasets_and_metrics}
        \subsubsection{Video evaluation} 
        \label{subsubsec:video_evaluation}
        In this part, we provide the dataset preparation and training details for video quality evaluation experiment.

        \noindent\textbf{Dataset Configuration.} We utilize the RT-1 dataset~\cite{brohan2022rt}, comprising 80K video demonstrations spanning over 590 distinct manipulation tasks categorized as Picking, Placing, Knocking and Opening/Closing. The structured composition of action-object pairs enables seamless integration with our action tree framework, eliminating the need for additional linguistic processing. To evaluate generalization capabilities, we adopt a task-based split allocating over 90\% of episodes (denoted as seen tasks) for training while reserving novel task combinations (unseen tasks) for testing.

        \noindent\textbf{Training Protocol.} Our implementation follows RoboDreamer's~\cite{zhou2024robodreamer} two-stage paradigm: initial training of a base video generation model at 64×64 resolution, followed by cascaded super-resolution (SR) diffusion for 256×256 output. This design strategically separates policy learning from visual refinement based on three considerations:
        \begin{itemize}
        \item \textbf{Decoupled Policy \& SR}: Empirical analysis reveals that action semantics predominantly emerge in low-resolution generation, enabling policy-agnostic SR processing
        \item \textbf{Training Efficiency}: High-resolution training demands much more computation per iteration and will be much slower considering our training budget
        \item \textbf{Metric Reliability}: The cascaded super-resolution module could affect video quality, causing uncertainty in the results of video quality related metrics
        \end{itemize}
        Nevertheless, we also train super-resolution networks to up-sample the video to $256 \times 256$ resolution.
        Detailed generation samples are shown in the appendix.

        \noindent\textbf{Metrics.} We mainly evaluate our method on two categories of metrics: image / video quality and task completion success rate.
        We use FID score \cite{heusel2017gans}, SSIM score \cite{wang2004image}, PSNR score,
        LPIPS score \cite{zhang2018unreasonable} and flow error score \cite{unterthiner2018towards} to evaluate image / video quality.
        When evaluating the image/video quality, we randomly selected 20 seen tasks and 20 unseen tasks, each task is evaluated for 50 initial states. So there will be 1000 generated videos in total for both seen and unseen setting.

        \subsubsection{Robot Planning}
        \label{subsubsection:robot_planning}

        \noindent \textbf{Planning tasks.} We select the six tasks of rlbench reported in RoboDreamer: Close box, Lamp Off, Lamp On, Stack Blocks, Lift Block and Take Shoes. Both short term tasks like closing the box, turn on the lamp and long term tasks like stack blocks and take shoes out of box.
        
        \noindent \textbf{Trainging details.} Following former works~\cite{zhou2024robodreamer,guhur2023instruction}, We first pick out the macro steps according to velocity and gripper status change. The macro step dataset contains mainly the key decision frames, making sure the model concentrates on the policy. We finetune the two generative model on the RT-1 pretrained version to utilize the prior knowledge and fostering the training process.

\begin{figure*}[!ht]
  \centering
  \includegraphics[width=\textwidth]{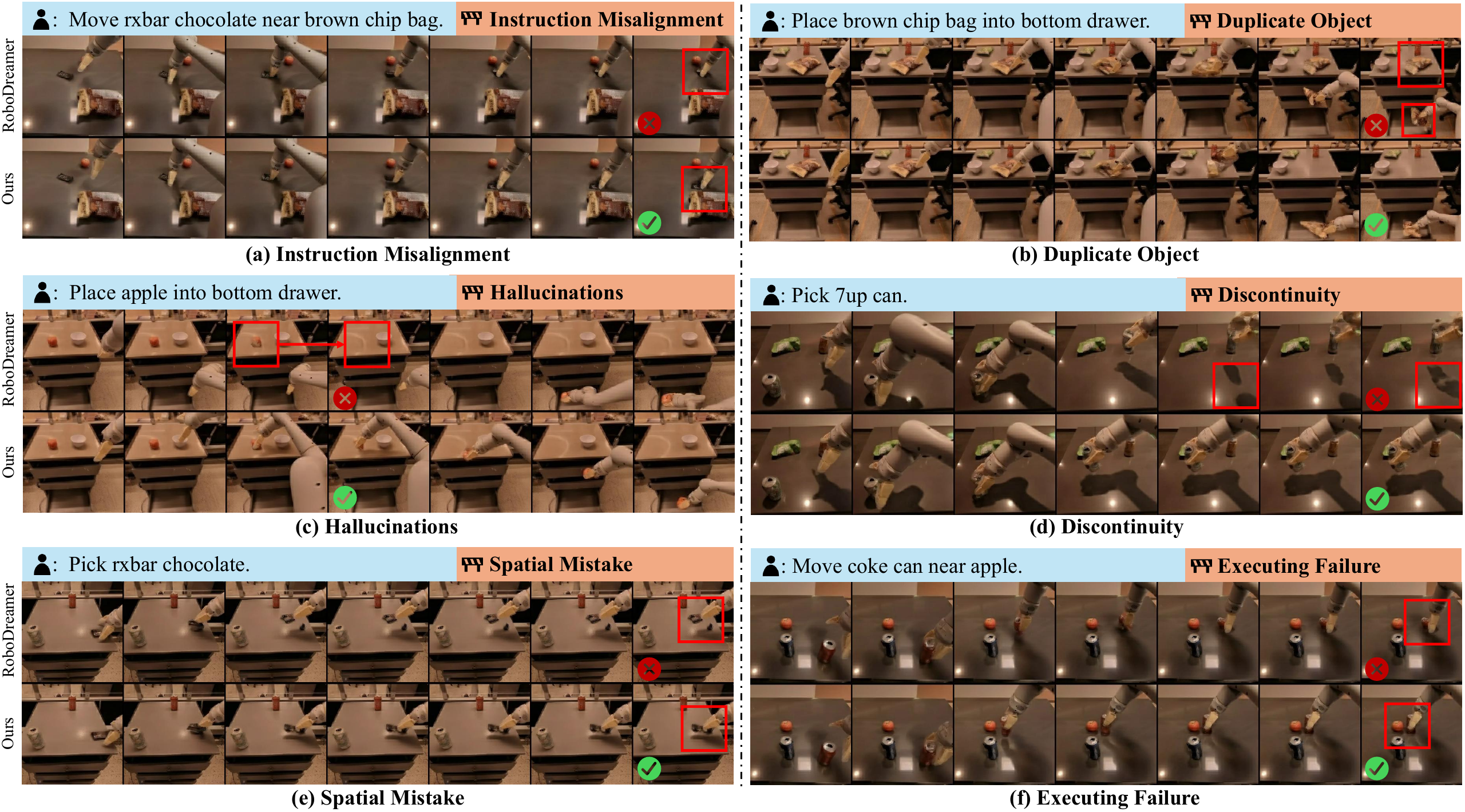}
  \vspace{-6mm}
  \caption{
    ManipDreamer alleviates multiple common defects in robotic video generation, including instruction misalignment, hallucinations, spatial errors, duplicate objects, temporal discontinuities, and failed executions.  
    This figure presents sample comparisons between RoboDreamer and ManipDreamer, illustrating the effectiveness of our proposed approach.
  }
  \label{fig:challenges}
\end{figure*}

\begin{table*}[t]
    \centering
    \caption{Success rate comparison on 6 RLBench tasks. Better results are in bold.}
    \vspace{-2mm}
    \label{tab:task_completion_rate}
    \resizebox{0.8\textwidth}{!}{
    \begin{tabular}{c|ccccccc}
        \toprule
        \textbf{Model} 
        & \textbf{Close Box} & \textbf{Lamp Off} & \textbf{Lamp On} 
        & \textbf{Stack Blocks} & \textbf{Lift Block} & \textbf{Take Shoes} 
        & \textbf{Average} \\
        \midrule
        RoboDreamer & 0.95 & \textbf{0.92} & 0.49 & 0.12 & 0.31 & 0.09 & 0.480\\
        ManipDreamer & \textbf{0.96} & 0.91 & \textbf{0.51} & \textbf{0.18} & \textbf{0.33} & \textbf{0.14} & \textbf{0.503}\\
        \bottomrule
    \end{tabular}
    }
\end{table*}
    
    \subsection{Main Results}
    \label{subsec:results}
        \subsubsection{Action Tree Effectiveness}
        \label{action_tree_effectiveness}
        To show the effectiveness of our proposed action tree method. We applied the action tree method on RoboDreamer, and removed the linguistic decomposition mechanism in it to compare these two action generalize methods.
        Tab.~\ref{tab:action_tree_exp} demonstrates our action tree's superiority over linguistic decomposition, achieving consistent improvements across 9/10 metrics. Notably, it reduces LPIPS by 4.9\% (0.071 vs 0.075) on seen tasks while maintaining comparable unseen task performance, validating its enhanced generalization through structured action representation, even though fewer forward steps are performed.
        
        \subsubsection{Visual Guidance Analysis}
        \label{visual_guidance_analysis}
        To illustrate our proposed method is more powerful than simply applying ControlNet in robotic video generation,
        we train four RoboDreamer ControlNet variants with different condition modalities: depth, semantic, RGB and dynamic mask as baselines and compare the video quality metric with our visual guiding method, the result are listed in Tab.~\ref{tab:video_quality}.
        
        \noindent \textbf{Sigal-modal ControlNets.} As evidenced in Tab.~\ref{tab:video_quality}, ControlNet variants demonstrate limited improvements over baseline RoboDreamer. For seen tasks, SAM-based conditioning achieves marginal gains of 0.75 percentage points in SSIM (0.7375 vs 0.7300) and 0.15 dB in PSNR (19.81 vs 19.66). Similar patterns emerge in LPIPS (2.05\% relative reduction) and flow error (6.7\% decrease from 3.800 to 3.545). Parallel trends hold for unseen tasks. This narrow performance improvement suggests single-modality conditioning inadequately addresses the complex spatial-temporal constraints in robotic video generation.
        
        \noindent \textbf{Multi-modal Control Adapter.} In contrast, our multi-modal control adapter (MCA) significantly outperforms single-modality ControlNet variants. On seen tasks, MCA achieves 21.06 PSNR (+1.32dB) with 0.0546 LPIPS (24.8\% reduction) over best ControlNet, confirming that synergistic multi-modal fusion captures richer spatial-temporal constraints than individual conditioning. On the unseen tasks, our MCA also get the best performance over all metrics except from FID, we conclude that our novel visual conditioning method largely improved the visual quality of generated robotic videos.

        \begin{table*}[htbp]
            \centering
            \caption{Visual Quality Metrics of Ablation study experiments. The best results are in bold in research question 1 and research question 3. The results with most decline from original Additive ManipDreamer are in italics in research question 2.Semantic stands for SAM feature and Mask stands for dynamic mask.}
            \label{tab:ablation_study_quality}
            \resizebox{\linewidth}{!}{
                \begin{tabular}{c|c|ccccc|ccccc}
                    \toprule
                    &\multirow{2}{*}{\textbf{Method}} & \multicolumn{5}{c|}{\textbf{Seen Tasks}} & \multicolumn{5}{c}{\textbf{Unseen Tasks}} \\
                    && \textbf{FID} $\downarrow$ & \textbf{SSIM} $\uparrow$ & \textbf{PSNR} $\uparrow$ & \textbf{LPIPS} $\downarrow$ & \textbf{Flow Error} $\downarrow$ 
                     & \textbf{FID} $\downarrow$ & \textbf{SSIM} $\uparrow$ & \textbf{PSNR} $\uparrow$ & \textbf{LPIPS} $\downarrow$ & \textbf{Flow Error} $\downarrow$\\
                    \midrule
                    \multirow{2}{*}{\textbf{RQ1}}&ManipDreamer (v, Additive Fusion)
                    & 8.181 & 0.7853 & 21.06 & 0.05702 & \textbf{3.049}
                    & 9.480 & 0.7982 & 21.05 & 0.05768 & \textbf{3.021}\\
                    &ManipDreamer (v, Cross Attention Fusion)
                    & \textbf{7.226} & \textbf{0.7922} & \textbf{21.27} & \textbf{0.05302} & 3.166
                    & \textbf{8.284} & \textbf{0.8024} & \textbf{21.20} & \textbf{0.05468} & 3.1650 \\
                    \midrule
                    \multirow{2}{*}{\textbf{RQ2}}&ManipDreamer (v, lower)
                    & \textbf{8.281} & 0.7711 & 20.50 & 0.06197 &  3.309
                    & \textbf{9.113} & 0.7797 & 20.61 & 0.05979 & 3.659 \\
                    &ManipDreamer (v, upper)
                    & 8.501 & \textbf{0.7899} & \textbf{20.97} & \textbf{0.05593} &  \textbf{3.097}
                    & 9.372 & \textbf{0.7959} & \textbf{21.06} & \textbf{0.05383} & \textbf{3.387} \\
                    \bottomrule
                \end{tabular}
            }
        \end{table*}

        \noindent\textbf{Visualization Analysis.} Qualitative results from our inference experiments provide compelling evidence for the system's improvements. As illustrated in Fig.~\ref{fig:challenges}, we systematically categorize six distinct failure modes prevalent in current world model's capability, each alleviated by our method:
        \textbf{Instruction Misalignment.} This critical failure occurs when generated actions contradict textual specifications. In Fig.~\ref{fig:challenges}(a), RoboDreamer erroneously positions the rxbar chocolate near an apple instead of the instructed brown chip bag, whereas our method demonstrates precise instruction grounding through action tree parsing.
        \textbf{Object Duplication.} Manifested as persistent object presence post-manipulation, this artifact violates object permanence principles. Fig.~\ref{fig:challenges}(b) shows RoboDreamer's output where the brown chip bag remains on the counter despite being supposedly placed in the drawer. Our spatial-aware generation eliminates such phantom objects in this case.
        \textbf{Physical Hallucinations.} Characterized by non-physical object interactions, these errors include teleportation and morphing artifacts. The baseline output in Fig.~\ref{fig:challenges}(c) demonstrates matter discontinuity where the target object instantaneously appears in the gripper, while our multi-modal constraints enforce coherent material transitions.
        \textbf{Temporal-Spatial Discontinuity.} We distinguish two subtypes: (1) Temporal discontinuity featuring implausible motion trajectories and (2) Spatial discontinuity with fractured object geometries, exemplified in Fig.~\ref{fig:challenges}(d) where the robotic arm's shadow exhibits abrupt fragmentation during movement;
        \textbf{Spatial Misunderstanding.} Distinct from discontinuity errors, this category involves fundamental 3D relationship misinterpretations. Fig.~\ref{fig:challenges}(e) exposes one RoboDreamer's failure to update spatial states - the rxbar chocolate appears both grasped and on the counter. Our depth-conditioned generation resolves this paradoxes.
        \textbf{Execution Failure.} Error mode includes premature termination and non-functional movements. As shown in Fig.~\ref{fig:challenges}(f), the baseline model aborts the action mid-execution, while our policy-aligned generation maintains complete action sequences.

        \subsubsection{Planning Ability}
        \label{planning_ability}
        
        Table~\ref{tab:task_completion_rate} demonstrates ManipDreamer's competitive task completion rates, outperforming RoboDreamer on 5 of 6 RLBench tasks. Key improvements include Stack Blocks (+33.3\% relative) and Take Shoes (+55.6\% relative), highlighting our method's advantage in complex, multi-step tasks. The 5.2\% average gain stems from better action decomposition and spatial-temporal modeling through our novel multi-modal conditioning framework and action tree method.

    \subsection{Ablation Study}
    \label{subsec:ablation_study}

        We conduct systematic ablation experiments to investigate three core research questions regarding our framework's design choices:
        
        
        \noindent \textbf{RQ1: Fusion Strategy.} What is the better way to incorporate fused features into the generative model?
        
        \noindent \textbf{RQ2: Layer Injection.} What is the difference in impact caused by guidance at different decode layers?
        
        
        For the first question, we compare the results of the additive fusion and cross attention fusion methods, in addictive fusion setting, conditions are added onto the hidden states when sequentially pass the 12 decoders, along with the residuals from corresponding encoder layer. Whilst in cross attention fusion methods, the hidden states are view as query and conditioning features are view as key and value when doing a cross attention. We found this process memory-consuming thus only one cross attention layer is used. We inject the conditioning features every 3 layers.
        As shown in Tab.~\ref{tab:ablation_study_quality}(RQ1), the cross attention fusion method results in better performance than additive fusion, achieving a margin of 0.87\% on SSIM by and 0.21 dB on PSNR in seen tasks. This performance advantage is consistently maintained for unseen tasks. The enhanced performance stems from the cross-attention mechanism's ability to effectively model complex spatiotemporal relationships between video dynamics and conditional features. 
        
        For the second question, we split the decoder layers of Denoising UNet into two groups: the upper 6 layers is close to middle block of UNet, where hidden states has smaller spatial size and larger hidden size and the lower 6 layers in the opposite, note we inject conditions every 3 layers, therefore there are two conditioned layers in each setting. We trained two variant of multi-modal control adapter applied on RoboDreamer, one only injecting the corresponding conditioning features into upper 6 layers, and the other at lower 6 layers in an additive fusion manner.
        As shown in Tab~\ref{tab:ablation_study_quality}(RQ2), upper-layer guidance yields better motion consistency (flow error 3.097 vs 3.309), while lower-layer conditioning enhances visual details (FID 8.28 vs 8.50). This aligns with the hierarchical nature of diffusion models - global structure forms early, local details refine later.

    Despite its higher memory footprint, cross-attention achieves superior metrics (Tab.~\ref{tab:ablation_study_quality}(RQ2)), getting a margin of SSIM by 0.7\% and PSNR by 0.21dB over additive fusion. This validates its enhanced capacity for modeling complex spatial-dynamic relationships.

\vspace{-2mm}
\section{Conclusion}
\label{conclusion}


    In this work, we propose ManipDreamer, a novel world model for robotic manipulation video synthesis that integrates structured instruction representation and multi-modal visual guidance.
    To enhance instruction-following ability, we introduce an action tree representation that explicitly captures the structural relationships among instruction primitives.
    Each instruction is encoded by traversing the tree, and the resulting embeddings are used to condition the dynamics generation in the world model.
    To improve visual quality, we further design a visual guidance adapter that fuses depth, semantic information, RGB and dynamic masks, effectively enhancing temporal and physical consistency during video generation, resulting in higher quality metrics.
    Extensive experiments on RLBench benchmarks demonstrate that ManipDreamer significantly outperforms prior methods, achieving substantial gains in video manipulation task success rates.
    We believe our approach provides a promising direction for advancing robotic video generation and opens up new possibilities for integrating structured reasoning and multi-modal guidance in the robotic world model.

\bibliographystyle{ACM-Reference-Format}
\bibliography{sample-base}
\clearpage
\appendix
\section{Multi Scene Modality Acquisition}
Due to lack of extra modality data in the RT-1 dataset, we obtain the depth, segmentation feature and dynamic mask using the following methods.

\noindent \textbf{Depth estimation acquisition.} We obtain depth using the estimation of Depth Anything v2, which is able to generate precise and consistent depth giving a single monocular image. In particular, we use the ViT-L version of Depth Anything v2.

\noindent \textbf{Segmentation feature acquisition.} We utilize the SAM2 model to get segmentation feature of the input scene snapshot. In particular, we use the SAM2 tiny for to save memory and forward time, which still performs well despite the smallest version of the series. 

\noindent \textbf{Dynamic mask acquisition.} We use the algorithm \ref{alg:dynamic_mask_generation} to obtain dynamic mask from the pixel level feature of SAM2.
\begin{algorithm}[H]
    \caption{Dynamic Mask Generation}
    \label{alg:dynamic_mask_generation}
    \begin{algorithmic}[1]
        \Require{First frame $I_0 \in \mathbb{R}^{3 \times H \times W}$, Other frames $I_1, I_2, \ldots, I_T \in \mathbb{R}^{3 \times H \times W}$}
        \Ensure{Final dynamic mask $M \in \mathbb{R}^{H \times W}$}
        \State $F_0 \leftarrow \text{Extract feature vectors from } I_0$ \Comment{$F_0 \in \mathbb{R}^{C_{sam} \times H \times W}$}
        \For{$t = 1$ to $T$}
            \State $F_t \leftarrow \text{Extract feature vectors from } I_t$ \Comment{$F_t \in \mathbb{R}^{C_{sam} \times H \times W}$}
            \State $S_t \leftarrow \text{Compute pointwise similarity between } F_0 \text{ and } F_t$ \Comment{$S_t \in \mathbb{R}^{H \times W}$}
            \State $M_t \leftarrow 1 - S_t$ \Comment{$M_t \in \mathbb{R}^{H \times W}$}
        \EndFor
        \State $M \leftarrow \frac{1}{T} \sum_{t=1}^T M_t$ \Comment{Average over all frames}
        \State \Return{$M$}
    \end{algorithmic}
\end{algorithm}

\section{Training Protocol}
    \subsection{Video Quality evaluations on RT-1}
    We trained RoboDreamer \cite{zhou2024robodreamer} as our base model for 240k steps (global batch size 128) using DDP across 8 H800 GPUs (2 days total). The UNet backbone remains frozen in all experiments to ensure fair comparison and demonstrate the effectiveness of our proposed enhancements.
    
    \subsection{Task Execution in RLBench}
    We collected 200 episodes per evaluated task in RLbench \cite{james2020rlbench}, supplemented with additional episodes from other tasks to increase dataset diversity. All tasks were trained jointly to produce a single unified model. The diffusion model was initialized with RT-1 pretrained weights and fine-tuned for approximately 60k steps, focusing exclusively on macro-step generation.

\section{Modality Weight Analysis}

\begin{figure}
    \centering
    \includegraphics[width=\linewidth]{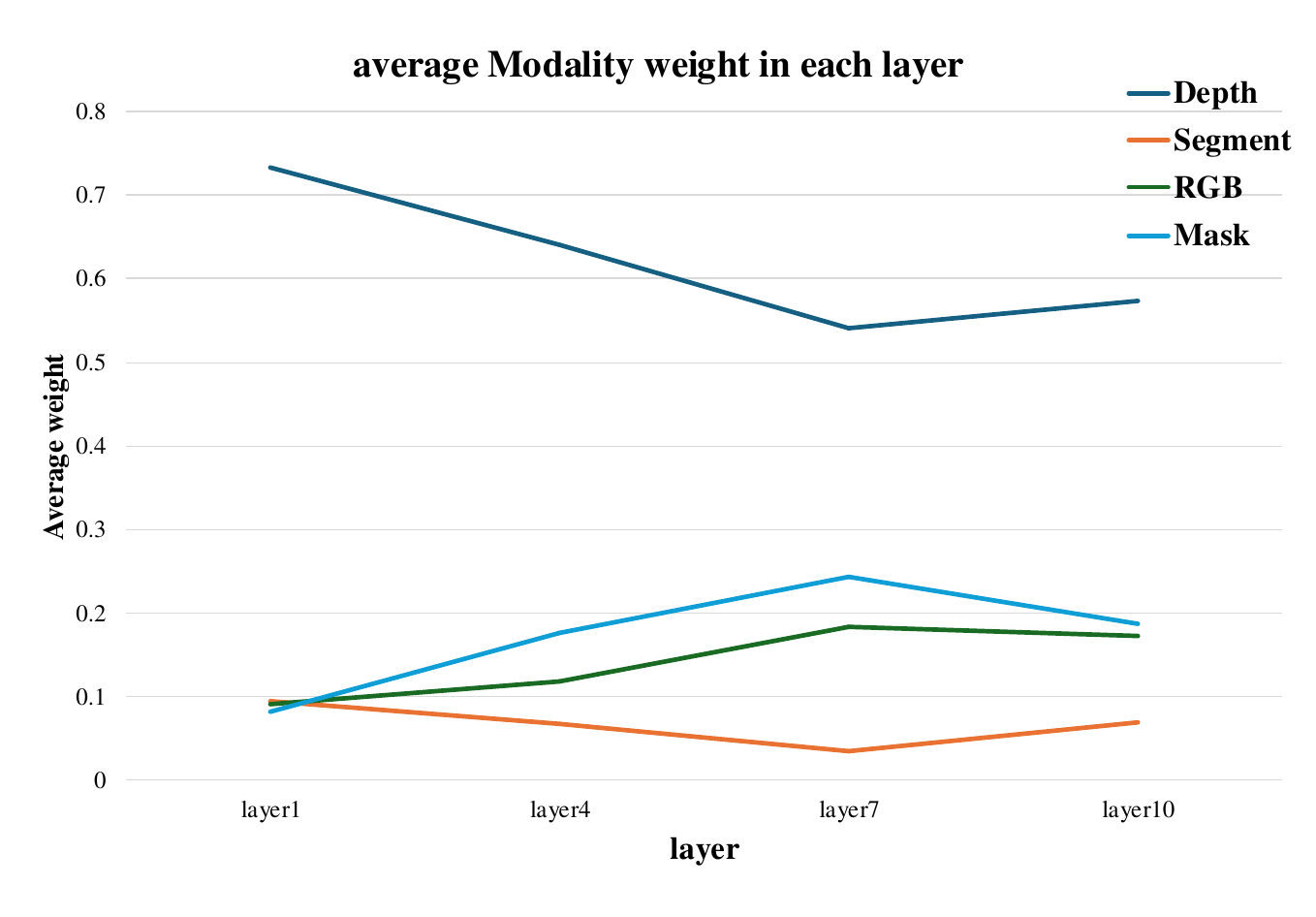}
    \caption{Distribution of average modality weights across layers 1, 4, 7, and 10 in our multi-modal visual adapter. The results demonstrate the relative importance of each modality in different network layers.}
    \label{fig:modality_weight}
    \end{figure}

    To quantify the contribution of each modality in our ManipDreamer framework, we analyzed the average weights generated by the patch-level router in our cross-attention fusion mechanism variant. We conducted 100 forward passes with initial observations and computed the mean weight values for each modality across different layers (1, 4, 7, and 10). As shown in Table~\ref{fig:modality_weight}, our analysis reveals that the model consistently assigns the highest weight to depth information across all layers, indicating its predominant role in the fusion process.
    
\end{document}